\documentclass[letterpaper]{article}
\usepackage{flairs}
\usepackage{times}
\usepackage{helvet}
\usepackage{courier}

\usepackage[font=footnotesize]{caption}
\usepackage{wrapfig}

\usepackage{natbib}  
\usepackage{caption}
\usepackage{bibentry}
\usepackage{algorithm}
\usepackage{amsmath}                                  
\usepackage{graphicx}
\usepackage{tikz}
\usetikzlibrary{arrows, automata}
\usetikzlibrary {arrows.meta}
\usepackage{import}
\usepackage{subcaption}
\usepackage{amsthm}
\usepackage[noend]{algpseudocode}
\usepackage{url}
\usepackage{array}
\usepackage{booktabs}
\usepackage{multirow}
\usepackage{amssymb}
\usepackage{float}
\newtheorem{theorem}{Theorem}
\newtheorem{proposition}[theorem]{Proposition}
\newtheorem{definition}[theorem]{Definition}
\newtheorem{corollary}[theorem]{Corollary}

\tikzset{
        circ/.style={shape=circle,draw,minimum size=1em}
}

\newcommand{\V}{{\mathbf V}}

\newcommand{\X}{{\mathbf X}}
\newcommand{\Y}{{\mathbf Y}}
\newcommand{\Z}{{\mathbf Z}}
\newcommand{\C}{{\mathbf C}}

\newcommand{\z}{{\mathbf z}}
\newcommand{\x}{{\mathbf x}}
\newcommand{\W}{{\mathbf W}}
\newcommand{\SSS}{{\mathbf S}}

\newcommand{\R}{{\mathbf C}}
\newcommand{\y}{{\mathbf y}}

\newcommand{\II}{{\mathcal I}}

\newcommand{\bedge}{\leftrightarrow}

\begin{document}
\title{Modeling and Discovering Direct Causes for Predictive Models}
\author{Yizuo Chen\textsuperscript{\rm 12}\thanks{This work was done during the author's internship at RTX Technology Research Center.},
Amit Bhatia\textsuperscript{\rm 2}\\
\textsuperscript{\rm 1}University of California, Los Angeles, USA\\
\textsuperscript{\rm 2}RTX Technology Research Center, Berkeley, USA\\
yizuo.chen@ucla.edu\ \ \ \ amit.bhatia2@rtx.com
}
\maketitle
\begin{abstract}
We introduce a causal modeling framework that captures the input-output behavior of predictive models (e.g., machine learning models). The framework enables us to identify features that \emph{directly cause} the predictions, which has broad implications for data collection and model evaluation. We then present sound and complete algorithms for discovering direct causes (from data) under some assumptions. Furthermore, we propose a novel independence rule that can be integrated with the algorithms to accelerate the discovery process, as we demonstrate both theoretically and empirically.
\end{abstract}

\section{Introduction}
Predictive models have become increasingly prevalent in decision-making over the past few decades. A predictive model predicts a set of \emph{outcomes} based on a set of input \emph{features}; see, e.g.,~\citep{MACKENZIE2013233,NeilsonIDT19,ELLIS2012261}. For instance, one may use a forecasting model to predict weather conditions based on data from the past week. Machine learning models are a common type of predictive models whose parameters are learned from data, e.g., support vector machines~\citep{ml/CortesV95}, decision trees~\citep{wa/BreimanFOS84}, and more recently, neural networks~\citep{Bishop1995,GoodBengCour16}. Other types of predictive models that do not involve machine learning include rule-based expert systems~\citep{rule-based-book} and probabilistic models constructed from domain knowledge~\citep{pearl1988-textbook,darwiche-textbook}.

\begin{figure}[tb]
\centering
\begin{subfigure}[b]{0.24\linewidth}
\centering
\includegraphics[width=0.8\linewidth]{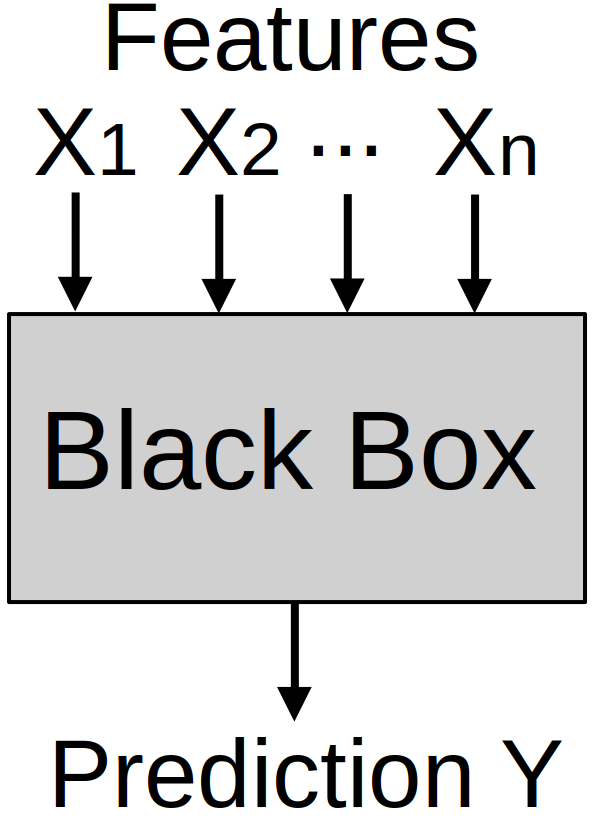}
\caption{setup}
\label{sfig:intro-eg0}
\end{subfigure}
\begin{subfigure}[b]{0.24\linewidth}
\centering
\begin{tikzpicture}[->,>=Stealth, scale=0.45, transform shape]
\node[font=\huge] (A) at  (0,0) {$A$};
\node[font=\huge] (D) at  (0,-1.5) {$D$};
\node[font=\huge] (S) at  (1.5,-1.5) {$S$};
\node[font=\huge] (P) at  (0.75,-3) {$P$};
\path (A) edge (D);
\path (D) edge (S);
\path (D) edge (P);
\path (S) edge (P);
\end{tikzpicture} 
\caption{$G$}
\label{sfig:intro-eg1}
\end{subfigure}
\begin{subfigure}[b]{0.24\linewidth}
\centering
\begin{tikzpicture}[->,>=Stealth, scale=0.45, transform shape]
\node[font=\huge] (A) at  (0,0) {$A$};
\node[font=\huge] (D) at  (1.5,0) {$D$};
\node[font=\huge] (P) at  (3,0) {$P$};
\node[font=\huge] (S) at  (1.5,-1.5) {$S$};
\path (A) edge (D);
\path (D) edge (P);
\path (D) edge (S);
\path (P) edge (S);
\end{tikzpicture} 
\caption{$G'$}
\label{sfig:intro-eg2}
\end{subfigure}
\begin{subfigure}[b]{0.24\linewidth}
\centering
\includegraphics[width=\linewidth]{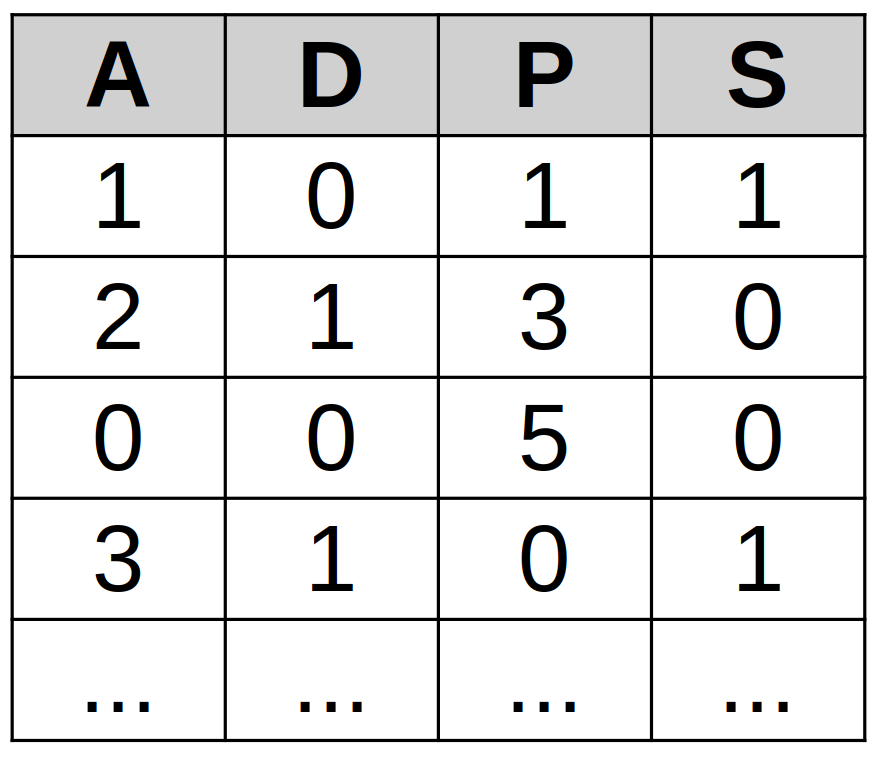}
\caption{data}
\label{sfig:fig1-pr}
\end{subfigure}
\caption{$G$ depicts the conventional causal graph over a patient's age ($A$), disease ($D$), symptom ($S$) and prescription ($P$), whereas $G'$ depicts the causal graph for the prediction of \(S\) from \(A, D, P.\)}
\label{fig:intro-eg}
\end{figure}

In this work, we consider a setup (in Figure~\ref{sfig:intro-eg0}) where the predictive models are treated as ``black boxes'' with configurations unknown to humans. This happens, for instance, when the model parameters are not publicly available or when the models (e.g., deep neural networks) are too complex to be transparent; see, e.g.,~\citep{cacm/Lipton18,Caruana15,Kohoutová20}.
To model the input-output behavior of predictive models under this setup, we introduce a class of causal graphs that represent predictive models using causal mechanisms.\footnote{The idea of treating machine learning models as causal mechanisms was mentioned briefly in~\citep{pods/Darwiche20}. In this work, we allow the causal mechanisms to exhibit uncertainties and consider the problem of discovering causal mechanisms from data.} This type of modeling appears to be different from the conventional approach yet effectively captures the data-generating process of the predictions. To illustrate, consider an example where a model is used to predict a patient's Symptom ($S$) based on their Age ($A$), Disease ($D$), and Prescription ($P$). Without bearing in mind that \(S\) is predicted from a model with inputs \(\{A, D, P\}\), one might construct the causal graph in Figure~\ref{sfig:intro-eg1} to model the interactions among variables. The graph, however, fails to capture the data generating process of \(S\), as illustrated by the mistaken conclusion that an intervention on \(P\) has no effect on the prediction for \(S.\) On the other hand, if we convert the predictive model into a causal mechanism for \(S,\) we attain the graph in Figure~\ref{sfig:intro-eg2} which correctly reveals the causal relations in this setup. As we will show later, all predictive models can be represented as causal graphs in this manner. This type of modeling is particularly useful when building causal graphs for large systems, where a predictive model, viewed as a system component, can be simply represented as a mechanism within the graph.

Once a causal graph is obtained, the \emph{direct causes} for predictions on the outcome \(Y\) become exactly the parents of \(Y\) in the graph. Identifying the direct causes for model predictions has a wide range of applications. First, it provides insights into which features contribute to the predictions, which has vast implications for model explainability and fairness; see, e.g.,~\citep{inffus/AliAEMACGSRH23,kdd/Ribeiro0G16,ecai/DarwicheH20,barocas-hardt-narayanan,aistats/ZafarVGG17}. Second, identifying features that do not directly cause the predictions allows us avoid unnecessary data collections in the future, which reduces the cost on data acquisition; see, e.g.,~\citep{jssam/smad007,Traskcost}. Our main question becomes: how can we discover these direct causes from data? To answer the question, we first propose two assumptions on the data distribution that ensure the direct causes are discoverable (uniquely determined). Under either assumption, the direct causes form a \emph{Markov boundary} of the outcome --- a notion introduced in~\citep{pearl1988-textbook} that has been studied extensively since then. By leveraging existing algorithms for discovering Markov boundaries, we develop \emph{sound and complete} methods for discovering direct causes. We show that one of the assumptions further simplifies the discovery process, leading to more efficient discovery algorithms. Another contribution of this work is the introduction of a novel independence rule, which, when integrated with existing algorithms, further accelerates the discovery process as we demonstrate both theoretically and empirically.

The paper is structured as follows. We start with some technical preliminaries in Section~2. In Section~3 we introduce the causal modeling for predictive models and formally define the notion of direct causes in the context. In Section 4 we propose two assumptions under which direct causes can be discovered from data, as well as algorithms for discovering the direct causes. We then show an independence rule that can be integrated into the algorithms to further improve the efficiency in Section~5. Section~6 presents empirical results that demonstrate the effectiveness of the independence rule. We close with some concluding remarks in Section~7.

\section{Technical Preliminaries}
\label{sec:preliminaries}
We assume all variables are discrete, though all the results can be extended to continuous domains. Single variables are denoted by uppercase letters (e.g., \(X\)) and their states are denoted by lowercase letters (e.g., \(x\)). Sets of variables are denoted by bold, uppercase letters (e.g., \(\X\)) and their instantiations are denoted by bold, lowercase letters (e.g., \(\x\)).

\subsection{Causal Models and Interventions}
We consider causal graphs in the form of acyclic directed mixed graphs (ADMGs)~\citep{richardson} as follows. 
\begin{definition}
\label{def:admg}
An \underline{acyclic directed mixed graph (ADMG)} is a graph that contains directed edges (\(\rightarrow\)) and bidirected edges (\(\bedge\)) and in which directed edges do not form any cycles.\footnote{Each \(X \bedge Y\) represents a hidden confounder \(U,\) i.e., \(X \leftarrow U \rightarrow Y.\) The class of ADMGs is more general than the class of DAGs. To illustrate, no DAG can capture the conditional independencies exhibited by the ADMG \(A \rightarrow B \bedge C \bedge D.\)}
\end{definition}
Figure~\ref{sfig:fig1-orig} depicts an ADMG over four variables. 
\begin{wrapfigure}[5]{r}{0.2\linewidth}
\vspace{-10pt}
\captionsetup{skip=0pt}
\begin{tikzpicture}[->,>=Stealth, scale=0.4, transform shape]
\node[font=\huge] (A) at  (0,0) {$A$};
\node[font=\huge] (B) at  (2,0) {$B$};
\node[font=\huge] (C) at  (4,0) {$C$};
\node[font=\huge] (Y) at  (2,-1.5) {$Y$};
\path[->] (A) edge (B);
\path[->] (B) edge[,<->,bend left=40] (C);
\path[->] (A) edge[,<->,bend left=60] (C);
\path[->] (B) edge (Y);
\path[->] (C) edge (Y);
\end{tikzpicture}
\caption{}
\label{sfig:fig1-orig}
\end{wrapfigure}
Let \(X\), \(Y\) be two variables in an ADMG, we say that \(X\) is a \emph{parent} of \(Y\), and \(Y\) a \emph{child} of \(X\) if \(X \rightarrow Y.\) Moreover, we say that \(X\) is an \emph{ancestor} of \(Y\), and \(Y\) a \emph{descendant} of \(X\) if there is a \emph{directed path} ($X \rightarrow \cdots \rightarrow Y$) from \(X\) to \(Y\). 
We say that \(X\) is a \emph{sibling} of \(Y\) if \(X \bedge Y\), and a \emph{spouse} of \(Y\) if \(X\) and \(Y\) share a same child. We say that \(X\) is a \emph{neighbor} of \(Y\) if it is a parent, child, or sibling of \(Y.\) A variable \(V\) is called a \emph{collider} on a path if \(\rightarrow V \leftarrow\), \(\bedge V \bedge\), \(\rightarrow V \bedge\), or \(\bedge V \leftarrow\) appears on the path and is called a \emph{non-collider} otherwise.

Intervention is a standard technique for studying the causal relations among events. By definition, an intervention fixes a variable to a specific state, which differs from naturally observing the state of a variable. For example, instructing (intervening) a patient to take a drug yields a different effect than seeing (observing) a patient taking a drug. We write \(do(X=x)\), or simply \(do(x)\), if an intervention fixes a variable \(X\) to the state \(x.\) Variable \(X\) exhibits a \emph{causal effect} on another variable \(Y\) if an intervention on \(X\) modifies the distribution of \(Y.\) This occurs only if \(X\) is an ancestor of \(Y\) in the causal graph~\citep{pearl09}.

\subsection{Independencies in Graphs and Distributions}
\label{sec:indep}
(Conditional) independence is a central notion in the domain of causal inference and discovery. In fact, the goal of causal discovery is to identify causal graphs consistent with the independencies encoded in a data distribution. We next review the definitions of independencies for both causal graphs and distributions and discuss the interplay between the two.

The independence relations in a causal graph (ADMG) are characterized by the notion of \emph{m-separation}~\citep{richardson}. By definition, let \(\X, \Y, \Z\) be three disjoint variables sets in an ADMG \(G,\) \(\X\) and \(\Y\) are said to be m-separated by \(\Z\), denoted \(\textnormal{msep}_G(\X,\Z,\Y)\), iff every path between \(\X\) and \(\Y\) satisfies the following: (1) a non-collider on the path is in \(\Z\); or (2) a collider on the path is not an ancestor of any variable in \(\Z\). In Figure~\ref{sfig:fig1-orig}, \(A\) and \(Y\) are m-separated by \(\{B,C\}\) but are not m-separated by \(\{B\}.\) 

Now consider a distribution \(\Pr\) over disjoint variable sets \(\X, \Y, \Z\). We say that \(\X\) and \(\Y\) are independent conditioned on \(\Z\) iff \(\Pr(\x | \y,\z) = \Pr(\x | \z)\) for all instantiations \(\x, \y, \z.\) Specifically, we write \(\II_{\Pr}(\X, \Z, \Y)\) if \(\X\) is independent of \(\Y\) given \(\Z\) and write \(\overline{\II_{\Pr}}(\X, \Z, \Y)\) otherwise~\citep{darwiche-textbook}. In practice, the distribution \(\Pr\) is typically represented by the empirical data as shown in Figure~\ref{sfig:fig1-pr}. Popular methods for testing independences from data include \(\chi^2\)-test~\citep{chi-square} and \(G\)-test~\citep{gtest}. These independence tests, however, have two bottlenecks as pointed out in~\citep[Ch.~5]{Spirtes-textbook}. The first is \emph{computational inefficiency} as the time required by each independence test is at least linear in the sample size. The second is \emph{sample inefficiency} as the number of samples required for stably testing \(\II_{\Pr}(\X, \Z, \Y)\) is exponential in the size of \(\Z.\)\footnote{To illustrate, suppose \(|\Z|=100\) and all variables in \(\Z\) are binary, there are \(2^{100}\) instantiations over \(\Z\) so we need at least \(2^{100}\) samples to ensure that each instantiation appears at least once.}

M-separations and independencies are related through the notions of independence map (I-MAP), dependency map (D-MAP), and perfect map (P-MAP)~\citep{pearl1988-textbook,darwiche-textbook}. We formally define these notions next.
\begin{definition}
\label{def:maps}
Let \(G\) be a causal graph and \(\Pr\) be a distribution over a same set of variables. We say that \(G\) is an \underline{I-MAP} of \(\Pr\) iff \(\textnormal{msep}_G(\X,\Z,\Y)\) implies \(\II_{\Pr}(\X, \Z, \Y)\) (for all \(\X,\Y,\Z\)); \(G\) is a \underline{D-MAP} of \(\Pr\) iff \(\II_{\Pr}(\X, \Z, \Y)\) implies \(\textnormal{msep}_G(\X,\Z,\Y)\); and \(G\) is a \underline{P-MAP} of \(\Pr\) iff \(G\) is both an I-MAP and a D-MAP of \(\Pr.\)
\end{definition}
We may sometimes say ``\(\Pr\) is an I-MAP of \(G\)'' to mean that ``\(G\) is an I-MAP of \(\Pr\)'', similarly for D-MAP and P-MAP. D-MAP is also called \emph{faithfulness} in the causal discovery literature. The notion of P-MAP is commonly required by existing causal discovery algorithms (such as PC~\citep{Spirtes-textbook}, FCI~\citep{Spirtes-textbook}, etc.) to ensure that the causal graph can be discovered from data. 

\subsection{Markov Boundary}
As we will discuss later, the discovery of direct causes for model predictions can be reduced to the discovery of Markov boundaries in some scenarios. Therefore, we also review the notion of Markov boundary along with some discovery algorithms here.
We start with the definition of Markov boundary in~\citep{pearl1988-textbook}.
\begin{definition}
\label{def:markov-boundary}
Let \(\Pr\) be a distribution over variables \(\X, Y.\) The \underline{Markov boundary} for \(Y\), denoted \(\textrm{MB}(Y),\) is the minimal subset of \(\X\) such that \(\II_{\Pr}(Y, \textrm{MB}(Y), \X \setminus \textrm{MB}(Y)).\)
\end{definition}
That is, \(Y\) is independent of other features when conditioned on its Markov boundary. Suppose a distribution \(\Pr\) is a P-MAP of some causal graph \(G\), then the Markov boundary of \(Y\) is unique and is equivalent to the \emph{Markov blanket} of \(Y\) in \(G.\) In particular, let the \emph{district} of \(Y\) be the variables connected to \(Y\) via bidirected paths (paths only involving bidirected edges), the Markov blanket of \(Y\) in an ADMG contains the following variables: the parents of \(Y\) ($\textsf{pa}(Y)$), the children of \(Y\) ($\textsf{ch}(Y)$), the spouses of $Y$ ($\textsf{sp}(Y)$), the district of $Y$ ($\textsf{dis}(Y)$), the parents of $\textsf{dis}(Y)$ ($\textsf{pa}(\textsf{dis}(Y))$), the districts of $\textsf{ch}(Y)$ ($\textsf{dis}(\textsf{ch}(Y))$), and the parents of $\textsf{dis}(\textsf{ch}(Y))$ ($\textsf{pa}(\textsf{dis}(\textsf{ch}(Y)))$)~\citep{tnn/YuLLC18}.\footnote{We can safely assume that the ADMGs are Maximal Ancestral Graphs (MAGs), a subtype of ADMGs that satisfy additional properties, since these two classes are Markov equivalent~\citep{Richardson-mag}. When \(G\) is a DAG, the Markov blanket contains the parents, children, and spouses of \(Y\)~\citep{pearl1988-textbook}.}

One key subroutine (procedure) widely used by existing Markov blanket discovery algorithms is \emph{adjacency search}, which identifies the neighbors of \(Y\) in the causal graph \(G\); see, e.g.,~\citep{kdd/TsamardinosAS03,amia/AliferisTS03,jmlr/AliferisSTMK10}. The procedure is based on the following observation: variables \(X, Y\) are adjacent to each other in \(G\) iff they are always dependent in \(\Pr\) regardless of the conditioned variables. To check whether there is an edge between two variables, the adjacency search enumerates all possible conditioned sets \(\Z \subseteq \X\) with an increasing size and removes a variable \(X\) from the neighbors of \(Y\) if \(\II_{\Pr}(X, \Z, Y).\)
Consider the causal graph \(G\) in Figure~\ref{sfig:fig1-orig} that is a P-MAP of some distribution \(\Pr.\) The adjacency search procedure initializes all features \(\{A,B,C\}\) to be the neighbors of \(Y.\) It then starts enumerating the conditioned sets \(\Z\) with an increasing size. When \(\Z = \{B,C\}\), it finds that \(\II_{\Pr}(A, \Z, Y)\) and therefore removes \(A\) from the neighbors of \(Y.\) The procedure finally concludes that the neighbors of \(Y\) are \(\{B,C\}\) after the enumeration of all feasible conditioned sets.

In the worst case, the number of independence tests required by adjacency search is exponential in the number of variables. One of the main focuses of this paper is to improve the efficiency of adjacency search, thereby accelerating the discovery of direct causes.

\section{Causal Modeling for Predictive Models}
We introduce a class of causal graphs called \emph{predictive graphs} to capture the input-output behavior of predictive models. Given a predictive model that takes a set of features \(\X\) and predicts an outcome \(Y,\) we construct a predictive graph that satisfies the following constraints: (1) \(Y\) cannot be a cause of any \(X \in \X\); and (2) there is no hidden confounder between a feature \(X\) and \(Y.\) These constraints follow naturally from the data generating process of \(Y\): intervening on predictions can never modify the input features, and the only possible causal factors for the predictions are the input features. We formally define the notion of predictive graphs.

\begin{definition}
\label{def:pred-graph}
Let \(\X\) be a set of features and \(Y\) be an outcome. A \underline{predictive graph} is an ADMG over \(\X, Y\) where the only possible edge between \(X \in \X\) and \(Y\) is \(X \rightarrow Y.\)
\end{definition}

We will use \(G(\X,Y)\) to denote a predictive graph wrt features \(\X\) and outcome \(Y.\) Figure~\ref{sfig:intro-eg2} depicts a predictive graph \(G(\{A, D, P\}, S).\) One key observation is that the predictive model is translated into the causal mechanism for \(Y\) in the predictive graph; that is, the causal mechanism (which involves \(Y\) and its parents) captures the input-output behavior of the predictive model. From now on, we shall assume that the data distribution \(\Pr(\X,Y)\) is always induced by some predictive graph \(G(\X,Y)\) in which the parents of \(Y\) correspond to the direct causes of the predictions for \(Y.\)\footnote{A distribution \(\Pr\) is said to be induced by a causal graph \(G\) iff it is attained by some parameterization of \(G.\) Moreover, \(G\) is guaranteed to be an I-MAP of the induced \(\Pr.\)}

In practice, however, predictive graphs are rarely available when predictive models are deemed black boxes. Hence, our goal is to discover the direct causes from data. This leads to two key questions: (1) when are the direct causes discoverable (uniquely determined)? (2) how can we identify these direct causes if they are indeed discoverable? Before addressing these questions, we formalize the definition of direct causes in~\citep{woodward} using interventions.

\begin{definition}
\label{def:direct-cause}
A variable \(X\) is a \underline{direct cause} of \(Y\) if \(\Pr(Y | do(x), do(\x')) \neq \Pr(Y | do(\x'))\) for some state \(x\) of \(X\) and instantiation \(\x'\) of \(\X \setminus \{X\}.\)
\end{definition}

That is, variable \(X\) is a direct cause of \(Y\) iff an intervention on \(X\) affects the distribution of \(Y\) while fixing the states of other variables. The definition suggests that discovering direct causes requires conducting interventions (experiments) and is impossible to infer from observational studies in general.\footnote{See~\citep{why-book} for a discussion on different layers of causal hierarchy.} However, under the assumption that the distribution is induced by some predictive graph, we can identify direct causes without the need of interventions as follows.

\begin{proposition}
\label{prop:direct-cause}
Let \(G(\X,Y)\) be a predictive graph that induces a distribution \(\Pr\) where \(\Pr(\X) > 0.\)\footnote{The positivity assumption ensures \(\Pr(Y | \X)\) is well-defined.} Then \(X \in \X\) is a direct cause of \(Y\) by Definition~\ref{def:direct-cause} iff \(\overline{\II_{\Pr}}(X, \X \setminus \{X\}, Y).\)
\end{proposition}

The proposition immediately suggests a naive method for discovering direct causes: check whether \(\overline{\II_{\Pr}}(X, \X \setminus \{X\}, Y)\) holds for each feature \(X.\) This method, however, is not sample-efficient since the set \(\X \setminus \{X\}\) may grow with the number of features; see our earlier discussion on sample-efficiency in Section~\ref{sec:indep}. We next propose some assumptions under which this issue can be mitigated.

\section{Assumptions for Discovering Direct Causes}
\label{sec:faithfulness}
We propose two assumptions under which the direct causes of the predictions are discoverable. In both cases, we show that the direct causes become equivalent to the Markov boundary (Definition~\ref{def:markov-boundary}) so we can leverage methods for discovering Markov boundaries for discovering direct causes.

\subsection{Canonicity}
We start with an assumption, which we call \emph{canonicity}, that is commonly assumed by existing algorithms to ensure that Markov blankets are discoverable.
\begin{definition}
\label{def:canonical}
A distribution \(\Pr\) is said to be \underline{canonical} if it is a P-MAP of some causal graph \(G.\)
\end{definition}

Note that the causal graph \(G\) in Definition~\ref{def:canonical} may be any ADMG, rather than a predictive graph, making the assumption quite general. The following result shows that direct causes are always discoverable for canonical distributions.
\begin{theorem}
\label{thm:canonical-mb}
If \(\Pr(\X, Y)\) is canonical, then the direct causes of \(Y\) form a unique Markov boundary of \(Y\) in \(\Pr.\)
\end{theorem}
That is, the problem of discovering direct causes in a predictive graph can be reduced to the problem of discovering the Markov blanket when the given distribution is canonical. Hence, we can leverage the existing methods for discovering Markov blankets under ADMGs such as the M3B algorithm~\citep{tnn/YuLLC18}.
To illustrate, suppose that a distribution \(\Pr\) is a P-MAP of the causal graph \(G\) in Figure~\ref{sfig:efficiency2}, then the direct causes of \(Y\) are exactly the Markov blanket of \(Y\) in \(G,\) which contains \(\{A, B, C, D, E, F\}.\)

\begin{figure}[tb]
\centering
\begin{subfigure}[b]{0.49\linewidth}
\centering
\begin{tikzpicture}[->,>=Stealth, scale=0.4, transform shape]
\node[font=\huge] (A) at  (0,0) {$A$};
\node[font=\huge] (Y) at  (0,-1.5) {$Y$};
\node[font=\huge] (C) at  (1.5,-1.5) {$C$};
\node[font=\huge] (D) at  (0,-3) {$D$};
\node[font=\huge] (H) at  (0,-4.5) {$H$};
\node[font=\huge] (E) at  (-2,-3) {$E$};
\node[font=\huge] (G) at  (-2,-4.5) {$G$};
\node[font=\huge] (F) at  (-4.5,-3) {$F$};
\node[font=\huge] (B) at  (-4.5,-1.5) {$B$};
\path[->] (A) edge (Y);
\path[->] (Y) edge (D);
\path[->] (C) edge (D);
\path[->] (D) edge (H);
\path[->] (D) edge[<->,bend right=30] (E);
\path[->] (E) edge (G);
\path[->] (E) edge[<->,bend right=30] (F);
\path[->] (B) edge (F);
\end{tikzpicture}
\caption{}
\label{sfig:efficiency2}
\end{subfigure}
\begin{subfigure}[b]{0.49\linewidth}
\centering
\begin{tikzpicture}[->,>=Stealth, scale=0.4, transform shape]
\node[font=\huge] (MA) at  (1.5,0) {$A$};
\node[font=\huge] (Y) at  (1.5,-1.5) {$Y$};
\node[font=\huge] (MB) at  (4.5,-1.5) {$B$};
\node[font=\huge] (MC) at  (3,-3) {$C$};
\path[->] (MA) edge (Y);
\path[->] (Y) edge (MC);
\path[->] (MB) edge (MC);
\end{tikzpicture}
\caption{}
\label{sfig:faithful2}
\end{subfigure}

\caption{Causal graphs to illustrate different assumptions.}
\label{fig:efficiency-example}
\end{figure}

\subsection{Weak Faithfulness}
\label{ssec:weak-faithful}
Our second assumption is a weaker type of faithfulness that imposes constraints on the distributions induced by the true predictive graph. As we will show later, the assumption not only makes the direct causes discoverable but also leads to an improvement on the computational efficiency.

\begin{definition}
\label{def:faithfulness}
A distribution \(\Pr(\X,Y)\) is \underline{weakly faithful} if \(X \in \X\) is a direct cause of \(Y\) only if \(\overline{\II_{\Pr}}(X, \Z, Y)\) for all \(\Z \subseteq \X \setminus \{X\}.\)
\end{definition}
Intuitively, weak faithfulness requires that \(Y\) always depends on the direct causes regardless of the conditioned set. This assumption is likely to hold, for instance, when the predictive model is a polynomial regression. 
To see when the assumption may be violated, let \(\Pr\) be a P-MAP of the causal graph in Figure~\ref{sfig:faithful2}. In this case, \(\Pr\) is not weakly faithful because \(\II_{\Pr}(Y, A, B)\), even though \(B\) is a direct cause of \(Y\) by Theorem~\ref{thm:canonical-mb}.
The following result shows that direct causes are always discoverable under weak faithfulness.

\begin{theorem}
\label{thm:markov-boundary}
If \(\Pr(\X, Y)\) is weakly faithful, then the direct causes of \(Y\) form a unique Markov boundary of \(Y\) in \(\Pr.\)
\end{theorem}

Another advantage of the weak faithfulness assumption is that it enables a faster discovery of direct causes compared to existing Markov blanket discovery algorithms for two reasons. First, the direct causes of \(Y\) coincide with the neighbors of \(Y\) (in the true predictive graph) under the weak faithfulness. Hence, all direct causes can be found through a single adjacency search for \(Y,\) avoiding the need for additional independence tests to discover non-neighbor variables (e.g., spouses) as required by Markov blanket discovery. Second, by Markov assumption, \(Y\) is independent of all other features when conditioned on the direct causes. This allows us skip the ``symmetry correction'' step in adjacency search;\footnote{Symmetry correction is essential for the correctness of Markov blanket discovery algorithms. To conclude that \(X\) is a neighbor of \(Y\), we need to further check that \(Y\) is adjacent to \(X\) in addition to checking that \(X\) is adjacent to \(Y\); see line~13-18 of Algorithm~\ref{alg:smart-search}.} see~\citep{ml/TsamardinosBA06} for more details.

Algorithm~\ref{alg:direct-causes} shows the details of adjacency search. Under the weak faithfulness assumption, the direct causes of model predictions can be discovered by calling \(\textsc{adj-search}\) in Algorithm~\ref{alg:direct-causes} while skipping lines~13-18.\footnote{The independence test \(\II_{\Pr}\) may be \(\chi^2\)-test (or $G$-test).}

Before moving on to show another technique for optimizing the discovery process, we note that a distribution \(\Pr\) can be both canonical and weakly faithful. This happens, for example, when \(\Pr\) is a P-MAP of a predictive graph. 

\section{Optimization with an Independence Rule} \label{sec:ind-rule}

We next introduce a novel independence rule that can be integrated into the adjacency search to accelerate the discovery process when \(\Pr\) is canonical. This result can be combined with the optimization technique mentioned in the previous section if \(\Pr\) is also weakly faithful. We start with the following theorem that introduces a key independence rule.

\begin{figure}[tb]
\centering
\begin{subfigure}[b]{0.49\linewidth}
\centering
\begin{tikzpicture}[->,>=Stealth, scale=0.4, transform shape]
\node[font=\huge] (E) at  (0,0) {$A$};
\node[font=\huge] (G) at  (1.5,0) {$B$};
\node[font=\huge] (S) at  (3,0) {$C$};
\node[font=\huge] (Y) at  (4.5,0) {$D$};
\node[font=\huge] (D) at  (2.25,-1.5) {$Y$};
\path[->] (E) edge (D);
\path[->] (G) edge (D);
\path[->] (S) edge (D);
\path[->] (Y) edge (D);
\path[->] (G) edge (S);
\path[->] (E) edge[bend left=30] (S);
\end{tikzpicture}
\caption{}
\label{sfig:fig2-g2}
\end{subfigure}
\begin{subfigure}[b]{0.49\linewidth}
\centering
\begin{tikzpicture}[->,>=Stealth, scale=0.4, transform shape]
\node[font=\huge] (A1) at  (0,0) {$A_1$};
\node[font=\huge] (B1) at  (1.5,0) {$B_1$};
\node[font=\huge] (A2) at  (3,0) {$A_2$};
\node[font=\huge] (ds) at  (4.5,0) {$\cdots$};
\node[font=\huge] (An) at  (6,0) {$A_n$};
\node[font=\huge] (Y) at  (3,-2) {$Y$};
\path[->] (A1) edge[<->,bend left=60] (B1);
\path[->] (B1) edge[<->,bend left=60] (A2);
\path[->] (A2) edge[<->,bend left=60] (ds);
\path[->] (ds) edge[<->,bend left=60] (An);
\path[->] (A1) edge (Y);
\path[->] (A2) edge (Y);
\path[->] (An) edge (Y);
\end{tikzpicture}
\caption{}
\label{sfig:efficiency1}
\end{subfigure}
 \caption{Examples of predictive graphs.}
 \label{fig:fig2}
\end{figure}

\begin{theorem}
\label{thm:indep}
Let \(\Pr\) be a distribution over disjoint variable sets \(\X, \Y, \Z, \W.\) If \(\II_{\Pr}(\X, \Z \cup \W, \Y)\) and \(\II_{\Pr}(\X \cup \Z, \emptyset, \W)\), then \(\II_{\Pr}(\X, \Z, \Y).\)
\end{theorem}

This result allows us to skip the independence test on \(\II_{\Pr}(\X, \Z \cup \W, \Y)\), which involves a larger conditioned set, if we know that \(\overline{\II_{\Pr}}(\X, \Z, \Y)\) and \(\II_{\Pr}(\X \cup \Z, \emptyset, \W),\) which involve smaller conditioned sets. This method can be applied widely to skip independence tests in adjacency search, where the independence tests are conducted with increasingly larger conditioned sets. Again, skipping independence tests speeds up the adjacency search and, consequently, the discovery of direct causes, for the complexity of discovery algorithms is dominated by the number independence tests.

We next define a notion that can be used to characterize the scenarios in which an independence test can be skipped.

\begin{definition}
\label{def:dep-neighbors}
A variable set \(\V\) is said to be \underline{I-decomposable} wrt distribution \(\Pr\) if \(\V\) can be partitioned into non-empty sets \(\V_1\) and \(\V_2\) where \(\II_{\Pr}(\V_1, \emptyset, \V_2).\)
\end{definition}

We can employ the notion of I-decomposability as follows.
Suppose we want to test \(\II_{\Pr}(X, \Z, Y)\), the classical method applies an independence test, which can be quite time consuming under large samples. On the other hand, suppose we know that \(\Z' = (\Z \cup \{X\})\) is I-decomposable, we can skip the independence test and immediately conclude that the independence does not hold for the following reason. Given that \(\Z'\) is I-decomposable, we can partition \(\Z'\) into independent sets \(\Z_1, \Z_2\) where \(X \in \Z_1.\) Since adjacency search checks independence with an increasing size of conditioned set, it must have already concluded \(\overline{\II_{\Pr}}(X, \Z_1 \setminus \{X\}, Y).\) This implies \(\overline{\II_{\Pr}}(X, \Z, Y)\) by Theorem~\ref{thm:indep}. To illustrate, consider a distribution \(\Pr\) that is a P-MAP of the predictive graph in Figure~\ref{sfig:fig2-g2}. During adjacency search, we can skip the independence test \(\II_{\Pr}(Y, \{A,D\}, B)\) since we already know \(\II_{\Pr}(B, \emptyset, \{A,D\})\) and \(\overline{\II_{\Pr}}(Y, \emptyset, B).\)
We call this optimization technique the \emph{I-decomposability rule} and insert it as a precondition to Algorithm~\ref{alg:direct-causes} line~7.
 
One practical question is how to efficiently check whether a set \(\V\) is I-decomposable. 
When \(\Pr\) is canonical, this can be done through the following procedure. Pick any \(V \in \V\) and initialize a set \(\SSS = \{V\}.\) Recursively add variables to \(\SSS\) as follows: for each \(V \in \V\) that is not in \(\SSS\), add \(V\) to \(\SSS\) if \(\overline{\II_{\Pr}}(V, \emptyset, \SSS).\) The set \(\V\) is I-decomposable iff \(\SSS \neq \X\) when no more variable can be added to \(\SSS\). We can avoid repeated independence tests by caching pairwise independencies.

\begin{algorithm}[tb]
\caption{Adjacency Search with Symmetry Correction}\label{alg:direct-causes}
\small
\begin{algorithmic}[1]
\Procedure{nonsym-search}{Features $\X$, Target $Y$, $\Pr$}
\State Initialize adjacent nodes \(\R \gets \X\)
\State depth \(d \gets 0\)
\While{\(d < |\R|\)}
\For{every \(W \in \R\)}
\For{every \(\Z \subseteq (\R \setminus \{W\})\) where \(|\Z|=d\)}
\If{\textit{\(\Z \cup \{W\}\) is I-decomposable}} continue
\EndIf
\If{\(\II_{\Pr}(Y, \Z, W)\)}  remove \(W\) from \(\R\)
\EndIf
\EndFor
\EndFor
\State \(d \gets d + 1\)
\EndWhile
\State \Return \(\R\)
\EndProcedure
\Procedure{adj-search}{Features $\X$, Outcome $Y$, $\Pr$}
\State $\textrm{neighbors}(Y) \gets \Call{nonsym-search}{\X, Y, \Pr}$
\State /* \textit{The following code is for symmetry correction} */
\For{every \(Z \in \textrm{neighbors}(Y)\)}
\State \(\W \gets \X \cup \{Y\} \setminus \{Z\}\)
\State $\textrm{neighbors}(Z) \gets \Call{nonsym-search}{\W, Z, \Pr}$
\If{\(Y \notin \textrm{neighbors}(Z)\)} 
\State $ \textrm{neighbors}(Y) \gets \textrm{neighbors}(Y) \setminus \{Z\}$
\EndIf
\EndFor
\State \Return \(\textrm{neighbors}(Y)\)
\EndProcedure
\end{algorithmic}
\label{alg:smart-search}
\end{algorithm}

The following theorem shows that the I-decomposability rule preserves the behavior of adjacency searches.
\begin{theorem}
\label{thm:soundness}
If \(\Pr(\X, Y)\) is a canonical distribution, then \(\textsc{adj-search}(\X,Y, \Pr)\) in Algorithm~\ref{alg:smart-search} yields the same result with or without line~7.
\end{theorem}
That is, we can integrate the I-decomposability rule into the Markov blanket discovery algorithms (such as M3B) while preserving their soundness and completeness. If the distribution \(\Pr\) is also weakly faithful, we can combine the I-decomposability rule with the results in Section~\ref{ssec:weak-faithful} to accelerate the discovery process to the maximum extent.
\begin{corollary}
\label{cor:soundness}
If \(\Pr(\X,Y)\) is canonical and weakly faithful, then \(\textsc{adj-search}(\X,Y,\Pr)\) in Algorithm~\ref{alg:smart-search} (with line~7 and without lines~14-18) yields the direct causes of \(Y.\)
\end{corollary}

We next briefly analyze the time complexity of the adjacency search. In particular, we focus on the number of independence tests required by the \textsc{nonsym-search} procedure since it is the dominating component of adjacency search (as shown in Algorithm~\ref{alg:direct-causes}). Similar to the result in~\citep{Spirtes-textbook}, the number of independence tests required by \textsc{nonsym-search} \emph{without the I-decomposability rule} is bounded by \(O(n \cdot \sum_{k=0}^{c}{n \choose k}),\) where \(n\) is the number of features and \(c = |\C|\) is the number of variables returned by the procedure. When we add the I-decomposability rule as a precondition in line~7, the procedure requires no more independence tests since more independence tests will be skipped.\footnote{The I-decomposability rule adds at most \(O(n^2)\) tests for pairwise independences. This overhead, however, is negligible when the causal graph is dense.} But how much speedup can the I-decomposability rule provide? The following result shows that the rule can sometimes reduce the number of independence tests \emph{exponentially}.
\begin{proposition}
\label{prop:fast-eg}
There exists a class of distributions \(\Pr\) with \(n\) features where \textsc{nonsym-search} with line~7 requires $O(n^3)$ independence tests while \textsc{nonsym-search} without line~7 requires $O(n \cdot \exp(n))$ independence tests.
\end{proposition}
The proof is based on constructing distributions that are P-MAP of the predictive graphs in Figure~\ref{sfig:efficiency1}. 

To summarize, we introduced two types of optimizations to speed up the discovery of direct causes. Both optimizations are based on improving the efficiency of the discovery of Markov boundaries. The first (Section~\ref{ssec:weak-faithful}) simplifies the discovery procedure when the distribution is weak faithful, while the second (Section~\ref{sec:ind-rule}) allows us to skip independence tests in adjacency search when the distribution is canonical.

Before presenting some empirical results, we note that \textsc{nonsym-search} in Algorithm~\ref{alg:direct-causes} is \emph{anytime}. Specifically, we can bound the depth \(d\) in Line~4 of Algorithm~\ref{alg:smart-search} without losing the true direct causes. This result is crucial for practitioners especially when computational resources are limited.

\section{Experiments}
\label{sec:exp}
We conduct experiments to further demonstrate the effectiveness of the I-decomposability rule. We compare the computational efficiency and sample efficiency of discovery algorithms with and without I-decomposability rule under the cases of (i) canonicity and weak faithfulness; and (ii) canonicity only. For case~(i), we compare the performance of six different algorithms: Algorithm~\ref{alg:smart-search} without line~7 (\textsc{adj}), Algorithm~\ref{alg:smart-search} with line~7 (\textsc{alg1}), Interleaved HITON-PC~\citep{amia/AliferisTS03,jmlr/AliferisSTMK10} (\textsc{i-hiton}), interleaved HITON-PC with the I-decomposability rule (\textsc{i-hiton-dec}), Semi-Interleaved HITON-PC~\citep{jmlr/AliferisSTMK10} (\textsc{si-hiton}) and Semi-Interleaved HITON-PC with the I-decomposability rule (\textsc{si-hiton-dec}). For case~(ii), we compare the performance of two algorithms: the M3B algorithm~\citep{tnn/YuLLC18} (\textsc{m3b}), and M3B algorithm with the I-decomposability rule (\textsc{m3b-dec}).\footnote{The I-decomposability rule can be incorporated into the HITON-PC algorithms and M3B algorithm, similar to Algorithm~\ref{alg:smart-search}, as a precondition for each independence test. We also implemented symmetry correction for the M3B algorithm.}

For all algorithms, we employ \(\chi^2\)-tests to test independences from data. When a discovery algorithm returns more direct causes than there actually are, we keep the direct causes that attain the lowest \(p\)-value among all independence tests conducted by the algorithm. In Algorithm~\ref{alg:direct-causes}, this can be implemented by recording the $p$-values for all independence tests in line~8.
In all experiments, we consider random causal models (Bayesian networks) that contain $100$ variables and are generated using the Erdős–Rényi method~\citep{erdos}. In case~(i), we generate random predictive graphs where the outcome variable has \(c\) parents.\footnote{We bound the maximal degree of features by $6$, where the degree of a node is defined as the number of its parents and children.} In case~(ii), we generate random ADMGs where the maximal degree of variables are bounded by \(d.\)

Our first set of experiments compares the computational efficiency of the algorithms. We consider causal graphs with different denseness by varying the number of direct causes \(c \in \{7,8,9,10\}\) in case~(i) and the maximal degree with \(d \in \{7,8,9,10\}\) in case~(ii). In both cases, the algorithms need to identify the direct causes from 100,000 instances randomly sampled from the true causal model. Table~\ref{tab:comp-eff1} records the average accuracy, runtime (in seconds), and number of independence tests (including those for checking I-decomposability) conducted by the algorithms over 20 runs. It is evident that algorithms with the I-decomposability rule attain shorter execution time and fewer independence tests than algorithms without the rule. In fact, the time was even halved by the I-decomposability rule in some cases, e.g., $c=10$ in case~(i). In general, the speedup is more significant in case~(i) than case~(ii). One possible explanation is that when \textsc{adj-search} (Algorithm~\ref{alg:smart-search}) is applied to outcome variables with more neighbors, more independence tests will be performed, leading to more being skipped by the I-decomposability rule, which outweighs the overhead of checking I-decomposability (shown in footnote~10). This is more likely to occur in case~(i) where the number of parents is fixed to large values.

We conduct further experiments to compare the sample efficiency of the algorithms. We vary the sample size from \(N \in \{1000,\) \(5000,\) \(10000,\) \(20000,\) \(50000,\) \(100000,\) \(150000,\) \(200000\}\) while fixing \(c=8\) in case~(i) and \(d=7\) in case~(ii). Figure~\ref{fig:exp} in the Appendix presents the accuracy achieved by different algorithms. It is clear that the algorithms with and without the I-decomposability rule achieve similar accuracy under all sample sizes. This suggests that the I-decomposability rule does not compromise the sample efficiency of existing algorithms.

\begin{table}[tb]
\centering
\scriptsize
\renewcommand{\arraystretch}{0.8}
\begin{tabular}{|c|c|c|c|c|c|}
\hline
{Methods} & {Metrics} & {$c=7$} & {$c=8$} & {$c=9$} & {$c=10$} \\
\hline
\multirow{3}{*}{\textsc{adj}} & Acc & 93.1 & 93.0 & 86.0 & 84.8\\
& Time & 3.1 & 4.1 & 5.3 & 6.5 \\
& \#CI & 2171 & 2853 & 3630 & 4309\\
\hline
\multirow{3}{*}{\textsc{alg1}} & Acc & \textbf{93.7} & \textbf{93.0} & \textbf{87.3} & \textbf{85.2}\\
& Time & \textbf{1.9} & \textbf{2.5} & \textbf{2.6} & \textbf{3.0}\\
& \#CI & \textbf{1497} & \textbf{1834} & \textbf{1923} & \textbf{2142}\\
\hline \hline
\multirow{3}{*}{\textsc{i-hiton}} & Acc & \textbf{96.6} & 95.0 & 89.8 & 89.0\\
& Time & 1.9 & 3.4 & 5.6 & 7.7\\
& \#CI & 1132 & 2061 & 3335 & 4477\\
\hline
\multirow{3}{*}{\textsc{i-hiton-dec}} & Acc & 96.3 & \textbf{95.0} & \textbf{90.0} & \textbf{90.2}\\
& Time & \textbf{1.2} & \textbf{1.9} & \textbf{3.0} & \textbf{3.5}\\
& \#CI & \textbf{685} & \textbf{1095} & \textbf{1650} & \textbf{1931}\\
\hline \hline
\multirow{3}{*}{\textsc{si-hiton}} & Acc & 96.0 & \textbf{95.0} & 90.0 & 88.4\\
& Time & 2.3 & 3.5 & 5.9 & 8.7\\
& \#CI & 1313 &  2062 & 3385 & 4887\\
\hline
\multirow{3}{*}{\textsc{si-hiton-dec}} & Acc & \textbf{96.0} & 94.8 & \textbf{90.2} & \textbf{89.6}\\
& Time & \textbf{1.4} & \textbf{2.0} & \textbf{2.8} & \textbf{3.8}\\
& \#CI & \textbf{805} & \textbf{1167} & \textbf{1560} & \textbf{2035}\\
\hline
\hline
{Methods} & {Metrics} & {$d=7$} & {$d=8$} & {$d=9$} & {$d=10$} \\
\hline
\multirow{3}{*}{\textsc{m3b}} & Acc & 86.9 & \textbf{73.8} & 74.7 & \textbf{71.3}\\
& Time & 52.1 & 178.6 & 818.5 & 1866.1\\
& \#CI & 48131 & 146322 & 523858 & 1090243\\
\hline
\multirow{3}{*}{\textsc{m3b-dec}} & Acc & \textbf{86.9} & 73.3 & \textbf{75.5} & 71.2\\
& Time & \textbf{41.9} & \textbf{156.5} & \textbf{794.6} & \textbf{1755.3}\\
& \#CI & \textbf{41865} & \textbf{129390} & \textbf{473593} & \textbf{1009157}\\
\hline
\end{tabular}
\caption{Average accuracy (Acc), time (Time), and number of independence tests (\#CI) for different methods. The I-decomposability rule is added to \textsc{alg1}, \textsc{i-hiton-dec}, \textsc{si-hiton-dec}, \textsc{m3b-dec}.}
\label{tab:comp-eff1}
\end{table}

\section{Conclusion}
We studied the problem of discovering features that directly cause the predictions made by predictive models, empowered by a causal modeling framework that represents the prediction process using causal graphs. We presented two assumptions under which the direct causes can be identified by leveraging existing methods for discovering Markov boundaries. Additionally, we proposed a novel independence rule that can be integrated with these algorithms to improve computational efficiency. This work demonstrates the application of causal tools to interpret predictive models, even in cases where the models are non-transparent, such as neural networks. Potential future works include identifying more conditions under which the direct causes can be efficiently discovered, studying the discovery of indirect causes for model predictions, and exploring the applications of the independence rule in broader contexts of causal discovery.

\section*{Acknowledgments}
This work is supported by RTX Technology Research Center. The authors would like to thank Adnan Darwiche (UCLA), and Isaac Cohen, Kishore Reddy, Adam Suarez, and Nathanial Hendler (RTX) for all the useful discussions and feedback.

\bibliography{flairs}
\newpage

\appendix
\section{Proofs}
\label{app:proof}
\subsection*{Proof of Proposition~\ref{prop:direct-cause}}
\begin{proof}
We first show the only-if direction. By contradiction, suppose \(\II_{\Pr}(Y, \X', X),\) then \(\Pr(y | x, \x') = C\) for all \(x\) where \(C\) is a constant. We can then compute \(\Pr(y | do(\x'))\) as follows.
\begin{equation*}
\begin{split}
&\Pr(y | do(\x')) = \sum_x \Pr(y | x, do(\x')) \Pr(x | do(\x')) \\
&= \sum_x \Pr(y | x, \x') \Pr(x | do(\x'))\;\;\; \text{(Rule~2 of do-calculus)}\\
&= C\sum_x \Pr(x | do(\x')) = C
\end{split}
\end{equation*}
Since \(\Pr(y | x, \x') = \Pr(y | do(x), do(\x'))\) by Rule~2 of do-calculus~\citep{pearl09}, we conclude \(\Pr(y | do(\x')) = \Pr(y | do(x), do(\x')) = C\) for all \(x,\) contradiction.

Now consider the if direction. Suppose \(\overline{\II_{\Pr}}(Y, \X', X),\) we can always find an instantiation \(y, \x'\) such that \(\Pr(y | x_1, \x') \neq \Pr(y | x_2, \x').\) Moreover, there must exists some state \(x^*\) that attains the largest \(\Pr(y | x^*, \x').\) Again, we can write out the \(\Pr(y | do(\x'))\) as follows
\begin{equation*}
\begin{split}
&\Pr(y | do(\x')) = \sum_x \Pr(y | x, do(\x')) \Pr(x | do(\x')) \\
&= \sum_x \Pr(y | x, \x') \Pr(x | do(\x'))\;\;\; \text{(Rule~2 of do-calculus)}\\
&< \sum_x \Pr(y | x^*, \x') \Pr(x | do(\x'))\\
&= \Pr(y | x^*, \x') \\
&= \Pr(y | do(x^*), do(\x'))\;\;\; \text{(Rule~2 of do-calculus)}\\
\end{split}
\end{equation*}
We conclude \(\Pr(y | do(\x')) \neq \Pr(y | do(x^*), do(\x')).\)
\end{proof}

\subsection*{Proof of Theorem~\ref{thm:canonical-mb}}
\begin{proof}
It suffices to check whether the result holds for the class of MAGs since every ADMG can be convert to some MAG that is Markov equivalent as shown in~\citep{richardson}.
As shown in~\citep{tnn/YuLLC18}, the minimal set that separates a target \(Y\) and other variables in a MAG is the Markov blanket (MB) of \(Y.\) We next show that \(\textrm{MB}(Y)\) are the only variables that satisfy the condition in Proposition~\ref{prop:direct-cause}. First, by weak union, \(\II_{\Pr}(Y, \X \setminus \{X\}, X)\) for all \(X \notin \textrm{MB}(Y)\) since \(\II_{\Pr}(Y, \X \setminus \textrm{MB}(Y), \textrm{MB}(Y))\) by the definition of Markov boundary. We next show that all \(X \in \textrm{MB}(Y)\) satisfies the condition. Note that \(\overline{\II_{\Pr}}(Y, \textrm{MB}(Y) \setminus X, X)\) for each \(X \in \textrm{MB}(Y)\). Otherwise, by contraction rule, \(\textrm{MB}'(Y) = \textrm{MB}(Y) \setminus \{X\}\) is also a valid Markov boundary, contradicting the uniqueness of MB. Moreover, for each \(X \in \textrm{MB}(Y)\), the active path from \(Y\) to \(X\) is still not m-separated even when we condition on more variables besides \(\textrm{MB}(Y).\) Hence, \(\overline{\II_{\Pr}}(Y, \X \setminus \{X\}, X).\)
\end{proof}

\subsection*{Proof of Theorem~\ref{thm:markov-boundary}}
\begin{proof}
Let \(\C\) be the direct causes of \(Y\) in \(G\) that satisfies the condition in Proposition~\ref{prop:direct-cause}. By m-separation, it is guaranteed that \(\II_{\Pr}(Y, \C,  \X \setminus \C),\) so \(\C\) is a valid Markov blanket. We are left to show that \(\C\) is unique and minimal. Suppose there exists another Markov boundary \(\W\) that omits some variable \(T \in \C,\) then \(\overline{\II_{\Pr}}(T, \W, Y)\) by the definition of weak faithfulness, contradicting \(\W\) being a Markov boundary.
\end{proof}

\subsection*{Proof of Theorem~\ref{thm:indep}}
\begin{proof}
First, by the rule of weak union, \(\II_{\Pr}(\X \cup \Z, \emptyset, \W)\) implies \(\II_{\Pr}(\X, \Z, \W).\) We then have the following:
\begin{equation}
\begin{split}
\Pr(\Y|\X,\Z) &= \sum_\W \Pr(\Y,\W|\X,\Z) \\
&= \sum_\W \Pr(\Y | \W, \X, \Z) \Pr(\W| \X, \Z) \\
&= \sum_\W \Pr(\Y | \W, \Z) \Pr(\W|\Z) \\
&= \Pr(\Y | \Z)
\end{split}
\end{equation}
which implies \(\II_{\Pr}(\X, \Z, \Y).\)
\end{proof}

\subsection*{Proof of Theorem~\ref{thm:soundness}}
\begin{proof}
It suffices to show that the output of \textsc{nonsym-search} is invariant with or without the I-decomposability rule. That is, whenever \(\Z \cup \{W\}\) is I-decomposable, it is guaranteed that \(\overline{\II_{Pr}}(Y, \Z, W).\) This follows from Theorem~\ref{thm:indep}. Suppose not, then \(\II_{Pr}(Y, \Z, W)\) together with \(\Z \cup \{W\}\) I-decomposable would imply that \(\II_{\Pr}(Y, \Z', W)\) for some \(\Z' \subset \Z.\) This leads to a contradiction since we should have removed \(W\) from \(\R\) much earlier.
\end{proof}

\subsection*{Proof of Proposition~\ref{prop:fast-eg}}
\begin{proof}
Consider the class of predictive graphs \(G\) shown in Figure~\ref{sfig:efficiency1} where \(n\) can be arbitrarily large. Let \(\Pr\) be the distributions that is a P-MAP of \(G\). Algorithm~\ref{alg:smart-search} with line~7 will first remove all \(B\)'s at \(d=2\) since \(\II_{\Pr}(B_i, \{A_i, A_{i+1}\}, Y).\) It takes \(O(n^3)\) conditional independence tests for \(d=2\) since we need to enumerate ${n \choose 2}$ conditioned variables for each of the \(n\) variables. No more conditional independence tests will be needed since any subsets of \(\{A_i\}_{i=1}^n\) with size greater than $2$ are I-decomposable.

We next consider the case without the I-decomposability rule. Similarly to the previous case, the adjacency search removes all \(B\)'s at \(d=2.\). However, the algorithm will continue searching for \(d=3, \dots, n-1\) afterward, taking a total of \(O(n\cdot \exp(n))\) independence tests.
\end{proof}

\begin{figure}[ht]
\centering
\begin{subfigure}[b]{0.99\linewidth}
\centering
\includegraphics[width=0.9\linewidth]{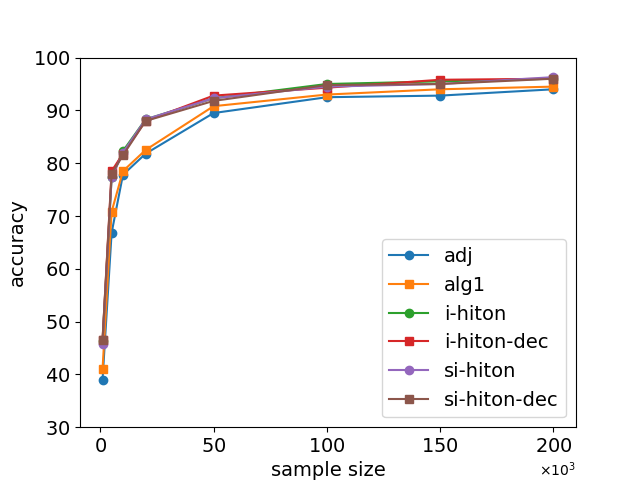}
\caption{canonicity \& weak faithfulness}
\label{sfig:exp-faithful}
\end{subfigure}
\begin{subfigure}[b]{0.99\linewidth}
\centering
\includegraphics[width=0.9\linewidth]{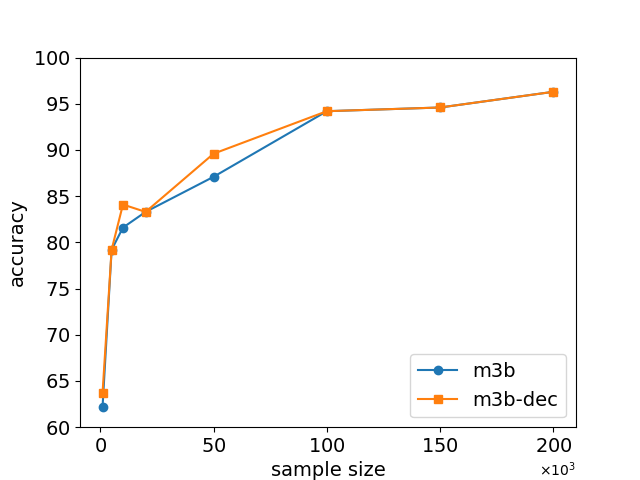}
\caption{canonicity only}
\label{sfig:exp-canonical}
\end{subfigure}
\caption{Accuracy of algorithms for identifying direct causes under various sample sizes.}
\label{fig:exp}
\end{figure}

\end{document}